\documentclass{article}  
\usepackage[margin=1.1in]{geometry}
\usepackage{graphicx}
\usepackage{subcaption}
\usepackage{amsmath}
\usepackage{amssymb}
\usepackage{float}
\usepackage{url}
\usepackage{hyperref}
\usepackage[utf8]{inputenc}
\usepackage{adjustbox}
\usepackage{array}
\graphicspath{ {images/} }

\newcommand{\nostarnote}[1]{}
\newcommand{\baad}[1]{} 

\title{\LARGE \bf
Robotic Detection of Marine Litter Using \\ Deep Visual Detection Models\footnote{This work is under review for ICRA 2019.}
}
\author{Michael Fulton$^{1}$, Jungseok Hong$^{2}$, Md Jahidul Islam$^{3}$, Junaed Sattar$^{4}$
	\thanks{The authors are with the Department of Computer Science \& Engineering, University of Minnesota, 200 Union St SE, Minneapolis, MN, 55455, USA
		{\tt\small \{$^{1}$fulto081, $^{2}$hong0442, $^{3}$islam034, $^{4}$junaed\} at umn.edu.}}
}
\begin{document}
\maketitle
\pagestyle{plain}

\begin{abstract}
Trash deposits in aquatic environments have a destructive effect on marine ecosystems and pose a long-term economic and environmental threat. Autonomous underwater vehicles (AUVs) could very well contribute to the solution of this problem by finding and eventually removing trash. 
This paper evaluates a number of deep-learning algorithms preforming the task of visually detecting trash in realistic underwater environments, with the eventual goal of exploration, mapping, and extraction of such debris by using AUVs. A large and publicly-available dataset of actual debris in open-water locations is annotated for training a number of convolutional neural network architectures for object detection. The trained networks are then evaluated on a set of images from other portions of that dataset, providing insight into approaches for developing the detection capabilities of an AUV for underwater trash removal. In addition, the evaluation is performed on three different platforms of varying processing power, which serves to assess these algorithms' fitness for real-time applications.

\end{abstract}

\section{Introduction}
\label{sec:introduction}

Marine debris poses a growing threat to the health of our planet. Beginning its life as discarded fishing gear, improperly recycled packaging, or simply discarded plastic grocery bags and soda bottles~\cite{epa1}, marine debris makes its way into the ocean or another body of water by a number of means and remains there~\cite{marinedebris}. There is virtually no place on earth that is unpolluted by marine debris, which kills and injures aquatic life, chokes ecosystems, and contaminates the water. Recycling and other efforts to keep debris out of the ocean have had limited impact. The vast amount of trash already in the ocean must somehow be removed. Despite the importance of this problem, there are few large-scale efforts attempting to combat it, due in part to the manpower required. We propose that a key element of an effective strategy for removing debris from marine environments is by using autonomous underwater vehicles (AUVs) to implement a trash detection and removal mechanism. 

In this paper, we examine the problem of detecting debris, particularly plastic debris, in an underwater environment, the first of a set of capabilities needed for such AUVs. We consider a number of deep learning-based visual object detection algorithms, build a dataset to train and evaluate them on, and compare their performance in this task by several metrics. Our core question is simply this: is deep-learning based visual detection of underwater trash plausible in real-time, and how do current methods preform in this area?

\begin{figure}[htp]
	\vspace{2.5mm}
	\centering
	\begin{subfigure}[b]{0.48\linewidth}
		\includegraphics[width=\linewidth]{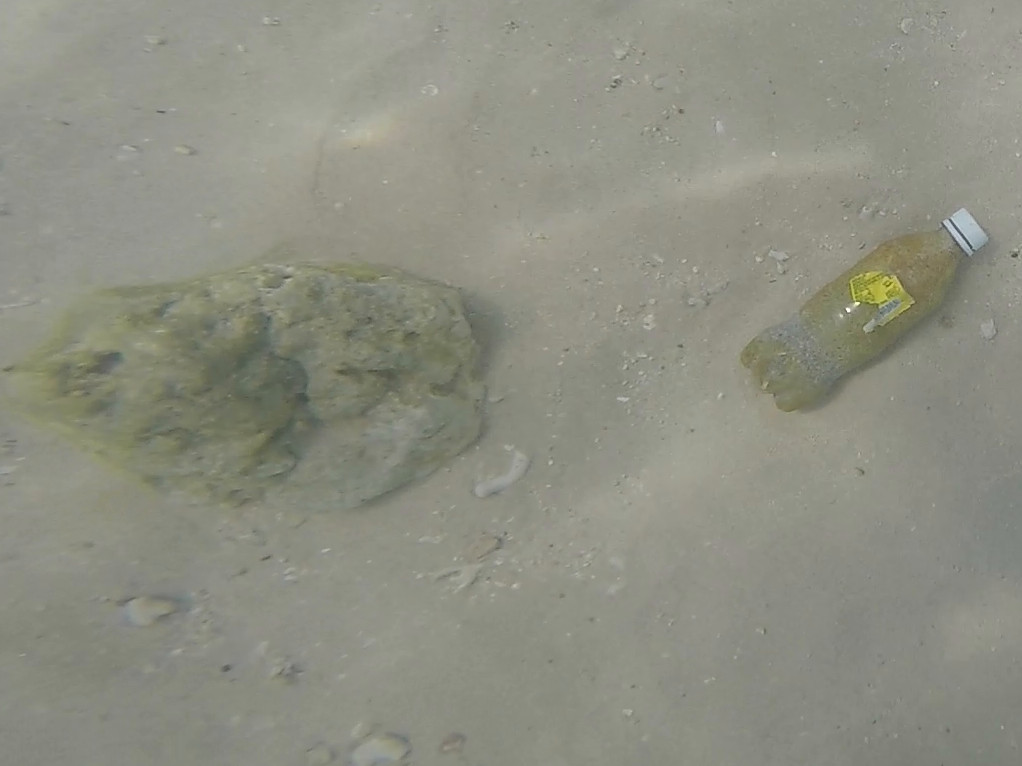}
		\caption{}
		\label{fig:trash_barbados}	
	\end{subfigure}\hspace{1mm}
	\begin{subfigure}[b]{0.48\linewidth}
		\includegraphics[width=\linewidth]{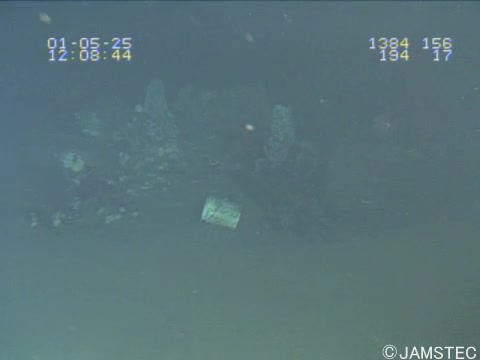}
		\caption{}
		\label{fig:trash_japan}
	\end{subfigure}
	\caption{Examples of plastic and other refuse material in various marine environments. ({\subref{fig:trash_barbados}}) A plastic bottle lying on the sea floor off the coast of Barbados; image collected by the authors, January 2018. ({\subref{fig:trash_japan}}) A beverage can on the sea bed off the coast of Japan, taken from the J-EDI dataset~\cite{JAMSTECDebri}.}
	\label{fig:trash_underwater}
\end{figure}

Detection of marine debris through purely visual means is a difficult problem to say the least. As with many visual object detection problems, small changes in the environment can cause massive changes in the appearance of an object, and nowhere is that more true than with underwater environments. Not only do light changes affect shallower water, but the changing turbidity (cloudiness) of the water may make the objects difficult or entirely impossible to detect~\cite{lu_underwaterimage_2016,fabbri_ehancing_2018}. Additionally, marine debris is rarely in pristine condition and degrades over time, so our detectors must be capable of recognizing different types of trash in any condition. Lastly, this problem is difficult to solve simply because of the enormous variety of objects that are considered marine debris. Even in the subgroup of plastic materials, the primary focus of this work and one of the largest groups of debris in the ocean, the variety is staggering. Figure~\ref{fig:trash_underwater} shows two examples of actual marine refuse having completely different appearances. The plastic bottle in Figure~\ref{fig:trash_barbados} is only one example of the many thousands of different styles of plastic bottles that may be found at the bottom of the ocean, to say nothing of plastic bags, containers, and other objects. As a starting point for the vast problem of detecting all marine litter, we focus on detecting plastic, one of the most prevalent and damaging types of littler in the ocean today.

To be useful for the goal of removing those plastics and other trash, these object detection algorithms must be able to run in near real-time on our robotic platform. To gauge their readiness for such a deployment, we tested all of our networks and models on three different devices, approximating the capabilities of an offline data processing machine, a high-powered robotic platform, and a low-powered platform.

In this paper, we make the following contributions.

\begin{itemize}
\item We evaluate the accuracy and performance of four state-of-the-art object detection algorithms for the problem of trash detection underwater. 
\item We evaluate the use of different training techniques on one particular network, YOLOv2.
\item We produce a unique dataset of marine debris for training deep visual detection models for this task.
\end{itemize}

\section{Related Work}
\label{sec:related}

Underwater autonomous robotics has been growing rapidly in significance and number of applications. Research has also focused on developing multi-robot systems for exploration of natural marine environments, both on the surface and underwater. Learning-based detection of underwater biological events for marine surveillance has also been explored~\cite{Bernstein2013IROS}. Additionally, underwater robots have been used in multi-robot systems for environmental surveillance (\emph{e.g.},~\cite{Girdhar2011MARE}), environmental mapping (\emph{e.g.},~\cite{kim2015active}), marine robotics for navigation and localization (\emph{e.g.},~\cite{s17051174,leonard2016autonomous,stutters2008navigation,paull2014auv,hidalgo2015review}), and more. 
There have been a number of efforts to identify the location of trash and marine litter and analyze their dispersal patterns in the open ocean, particularly in the presence of eddies and other strong ocean currents. Mace looked into at-sea detection of marine debris~\cite{mace_at-sea_2012} where he discussed possible applications of strategies and technologies. He points out that one of the major challenges is the debris being small, partially submerged, and buried in the sea bed. His focus is thus narrowed down to the floating debris on water parcel boundaries and eddy lines. Howell et al.~\cite{howell_north_2012} make a similar assessment, with their work investigating marine debris, particularly derelict fishing gear, in the North Pacific Ocean. Researchers have looked at the removal of debris from the ocean surface after the tsunami off the Japanese coast in 2011~\cite{dianna.parker_detecting_2015}. The work by Ge et al.~\cite{ge_semi-automatic_2016} uses LIDAR to find and map trash on beaches. Autonomous trash detection and pickup for terrestrial environments have also been investigated; for example, by Kulkarni et al.~\cite{Kulkarni2013CARE}. While this particular work is applied for indoor trash and uses ultrasonic sensors, a vision-based system can also be imagined.

Recent work by Valdenegro-Toro~\cite{Toro2016Trash} has looked into using forward-looking sonar (FLS) imagery to detect underwater debris by training a deep convolutional neural network (CNN) and has been demonstrated to work with approximately $80$ {\em per cent} accuracy. However, this work uses an in-house dataset constructed by placing objects that are commonly found with marine litter into a water tank and taking FLS captures in the tank. The evaluation was also performed on water tank data. This work demonstrates the applicability of CNNs and other deep models to the detection of small marine debris, but it is unknown whether it will be applicable to natural marine environments.

Some debris detection processes use remotely operated vehicles underwater (ROVs) to record sightings of debris and manually operate a gripper arm to remove items of interest.The FRED (Floating Robot for Eliminating Debris) vehicles have been proposed by Clear Blue Sea~\cite{clearBlueSea}, a non-profit organization for environmental protection and cleanup; however, the FRED platform is not an UAV. Another non-profit organization, the Rozalia project, has used underwater ROVs equipped with multibeam and side-scan sonars to locate trash in ocean environments~\cite{rozaliaROV}.

To enable visual or sensory detection of underwater trash using a deep-learned appearance model, a large annotated dataset of underwater debris is needed. For capturing the varied appearances across widespread geographical regions, this dataset needs to include data collected from a large spread of diverse underwater environments. Fortunately, some of these datasets exist, although most are not annotated for deep learning purposes. The Monterey Bay Aquarium Research Institute (MBARI) has collected a dataset over $22$ years to survey trash littered across the sea bed off the western seaboard of the United States of America~\cite{schlining_debris_2013}, specifically plastic and metal inside and around the undersea Monterey Canyon, which serves to trap and transport the debris in the deep oceans. Another such example is the work of the Global Oceanographic Data Center, part of the Japan Agency for Marine Earth Science and Technology (JAMSTEC). JAMSTEC has made a dataset of deep sea debris available online as part of the larger J-EDI (JAMSTEC E-Library of Deep-sea Images) dataset~\cite{JAMSTECDebri}. This dataset has images dating back to $1982$ and provides type-specific debris data in the form of short video clips. The work presented in this paper has benefited from annotating this data to create deep learning-based models.

\section{Data Source and Training Set Construction}
\label{sec:methodology}
This work evaluates four deep learning architectures for underwater trash detection. To evaluate these four networks, we construct a dataset for training and evaluation, define our data model, and annotate images for training.

\subsection{Dataset Construction}
The dataset for this work was sourced from the J-EDI dataset of marine debris which is described in detail in the previous section. The videos that comprise that dataset vary greatly in quality, depth, objects in scenes, and the cameras used. They contain images of many different types of marine debris, captured from real-world environments, giving us a variety of objects in different states of decay, occlusion, and overgrowth. Additionally, the clarity of the water and quality of the light vary significantly from video to video. This allows us to create a dataset for training which closely conforms to real-world conditions, unlike previous works, which mostly rely on internally generated datasets.

Our training data was drawn from videos labeled as containing debris, between the years of $2000$ and $2017$. From that portion of data, we further selected all videos which appeared to contain some kind of plastic. This was done in part to reduce the problem to a manageable size for our purposes, but also because plastic is an important type of marine debris~\cite{derraik2002pollution}. At this point, every video was sampled at a rate of three frames \textit{per} second to produce images which could be annotated to prepare them for use in learning models. This sampling produced over $240,000$ frames, which were searched for the best examples of plastic marine debris and then annotated. The final training dataset is composed of $5,720$ images, with dimensions of 480x320. 

\subsection{Data Model}
\label{subsec:detector}
Our trash detection data model has three classes, defined as follows: 
\begin{itemize}
\item \textbf{Plastic:} Marine debris, all plastic materials.
\item \textbf{ROV:} All man-made objects(\textit{i.e.}, ROV, permanent sensors, etc), intentionally placed in the environment.
\item \textbf{Bio:} All natural biological material, including fish, plants, and biological detritus.
\end{itemize}

This model is designed to find all plastic debris, regardless of their specific type. The reduced number of classes was chosen to improve inference speed as well as improve generality of the detector. The other classes in this model exist to avoid confusing plastic with intentionally placed man-made objects, plants, or animals. 

Other versions of this data model were tested: one including a label for timestamp text on the screen (rejected due to lower accuracy) and models with multiple classes for different plastic objects (rejected for lack of data on some classes and lower detection performance overall).  Both versions achieved lower accuracy than this model on all networks.

\begin{figure}
	\centering
		\includegraphics[width=0.5\linewidth]{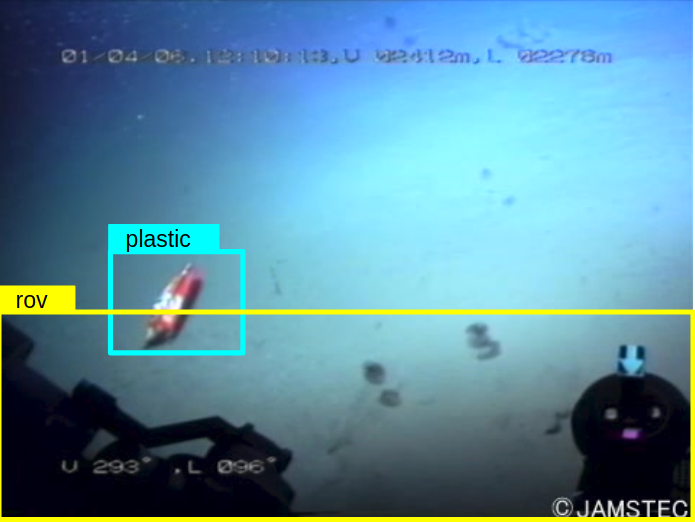}
		\caption{}
		\label{fig:2_label_one}
\caption{Example labeled image.}
\label{fig:one_label}

\end{figure}

Examples of labeled images from this data model can be seen in Figure \ref{fig:one_label}.

\subsection{Evaluation Data}
\label{subsec:evalData}
To evaluate the models, test images which contained examples of every class in our model were selected for a test set. However, these images were drawn from videos of objects which had not been used to train the network, so there should not be overlap between the training and test set) Multiple different types of plastic objects were selected, such as grocery bags, plastic bottles, etc. At least three different videos and a minimum of $20$ images per class were collected for each of these types of objects. In the end, $820$ such images were collected and annotated for testing. These images contained examples for each class, in a variety of environments which were intentionally selected to be challenging to the debris detectors, so as to provide a realistic evaluation of how these detectors would perform in field conditions.

\section{NETWORK ARCHITECTURES}
\label{sec:networks}
The four network architectures selected for this project were chosen from the most popular and successful object detection networks in use today. Each one has its own benefits and drawbacks, with varying levels of accuracy and runtime speeds. 
\subsection{YOLOv2}
You Only Look Once (YOLO) v2~\cite{redmon2016yolo9000} is the improved version of YOLO, an earlier network by the same authors. Although YOLO is much faster than Fast R-CNN, Faster R-CNN, and other models, it has notable localization errors and low recall in comparison with other state-of-the-art object detection algorithms. With several techniques such as batch normalization, a higher resolution for input images, and a change in the way bounding boxes are proposed to use anchor boxes, YOLOv2 improves upon the accuracy of its predecessor. In order to process faster, YOLOv2 uses a custom network called Darknet-19 rather than VGG-16, which is used by many object detection algorithms. Darknet-19 requires only about one sixth of the floating point operations of VGG-16 for a single pass of an input image.

\subsection{Tiny-YOLO}
Tiny YOLO~\cite{redmon2016you} was introduced along with YOLO. Tiny-YOLO is based on the neural network that has a smaller number of convolutional layers and filters inside the layers compared to the network used by YOLO while sharing the same parameters with YOLO for training and testing. In this way, tiny-YOLO runs at even faster speeds than YOLO while having similar mean Average Precision (mAP) values. Tiny-YOLO is especially useful to implement object detection for applications which have limited computational power such as field robotics. 

\subsection{Faster RCNN with Inception v2}
Faster RCNN~\cite{renNIPS15fasterrcnn} is an improvement on R-CNN~\cite{Girshick2014RCNN_CVPR} that introduces a Region Proposal Network (RPN) to make the whole object detection network end-to-end trainable. The RPN uses the last convolutional feature map to produce region proposals, which is then fed to the fully connected layers for the final detection. The original implementation uses VGG-16 \cite{simonyan2014very} for feature extraction; we use the Inception v2 \cite{szegedy2016rethinking} as feature extractor instead because it is known to produce better object detection performance in standard datasets \cite{tfzoo}.

\subsection{Single Shot MultiBox Detector (SSD) with MobileNet v2}
Single Shot MultiBox Detector (SSD)~\cite{liu2016ssd} is another object detection model that performs object localization and classification in a single forward pass of the network. The basic architectural difference of SSD with YOLO is that it introduces additional convolutional layers to the end of a base network which results in an improved performance. In our implementation, we use MobileNet v2 \cite{sandler2018inverted} as the base network. 

\section{TRAINING}
All four networks were trained according to the methods suggested by the community around the networks~\cite{tfzoo,alexeyab}, using on a Linux Machine, using four NVIDIA$^{TM}$ GTX 1080s.
In general, this involved fine-tuning the networks with some pretrained weights which had been trained on a well known object detection dataset. 
There were 5,720 images in our training set for each network, resized to 416x416 in the case of YOLOv2 and Tiny-YOLO.  For RCNN, SSD, and Tiny-YOLO, we simply fine-tuned each network for a few thousand iterations.

However, in the case of YOLO, the network was fine tuned, but also trained using a transfer learning technique, freezing all but the last several layers for training so that only the last few layers had their weights updated. In this way, we attempt to capitalize on the fact that earlier layers' pretrained weights already encode for basic image features, and simply train the last few layers of YOLOv2.
We report the results of these transfer-learned versions separately from the RCNN, SSD, and Tiny-YOLO results, to explore the possibility of using transfer learning to overcome a deficiency in dataset.  

\section{EVALUATION}
\label{sec:evaluation}
\subsection{Metrics}

\begin{table}[t]
  \centering
  \vspace{3mm}
  \begin{tabular}{l|c|>{\centering\arraybackslash}m{0.7cm}|*{3}{c}}
    Network& mAP&Avg. IoU&plastic AP&bio AP&rov AP \\
    \hline
    YOLOv2&47.9&54.7&\textbf{82.3}&9.5&52.1\\ 
    Tiny-YOLO&31.6&49.8&\textbf{70.3}&4.2&20.5\\ 
    Faster R-CNN&81.0&60.6&\textbf{83.3}&73.2&71.3\\
    SSD&67.4&53.0&\textbf{69.8}&6.2&55.9\\
  \end{tabular}
  \caption{Detection metrics in mAP, IoU, and AP.}
  \label{tab:detection_one}
\end{table}

\begin{table}
\centering
\begin{tabular}{  l | c | c | c }
Network & 1080 & TX2 & CPU \\
\hline
YOLOv2 &  74 & 6.2 & 0.11 \\
Tiny-YOLO  & \textbf{205} & \textbf{20.5} &  0.52 \\
Faster R-CNN & 18.75 & 5.66 & 0.97 \\
SSD & 25.2 & 11.25 & \textbf{3.19} \\
\end{tabular}
\label{tab:fps}
\caption{Performance metrics in frames per second.}
\vspace{-5mm}
\end{table}

The four different networks used for the marine debris detector were evaluate using the following metrics, for which we report two standard performance metrics.

\begin{itemize}
\item mAP (Mean Average Precision) is the average of precision at different recall values. Precision is the ratio $precision = \frac{TP}{TP+FP}$, and recall is the ratio $recall = \frac{TP}{TP+FN}$, where TP, FP, and FN stand for True Positive, False Positive, and False Negative.
\item IoU (Intersection Over Union) is a measure of how well predicted bounding boxes fit the location of an object, defined as $IoU=\frac{\text{area of intesection}}{\text{area of union}}$, where the intersection and union referred to are the intersection and union of the true and predicted bounding boxes.
\end{itemize}

These two metrics concisely describe the accuracy and quality of object detections. 

\subsection{Hardware}
Additionally, runtime performance metrics are provided on a GPU (NVIDIA\texttrademark{} 1080), embedded GPU (NVIDIA\texttrademark{} Jetson TX2), and CPU(an Intel\texttrademark{} i3-6100U) in terms of the number of frames which can be processed per second (FPS). The hardware used for evaluation is only relevant to the runtime.  Performance metrics, such as mAP, IoU, and recall, can be calculated identically on any device. The devices were selected to provide an indication of how these models could be expected to perform in an offline manner (GPU), in real-time on a computationally powerful robotic platform (embedded GPU), and in a lower-power robotic platform (CPU).

\begin{table}[t]
  \centering
  \vspace{3mm}
  \begin{tabular}{l|c|>{\centering\arraybackslash}m{0.7cm}|*{3}{c}}
    YOLO Training& mAP&Avg. IoU&plastic AP&bio AP&rov AP \\
    \hline
    Fine Tuned&47.9&54.7&\textbf{82.3}&9.5&52.1\\ 
    Last 4 Layers&33.9&45.4&\textbf{71.3}&13.6&17.0\\ 
    Last 3 Layers&39.5&34.1&\textbf{74.6}&19.9&23.9\\
  \end{tabular}
  \caption{Detection metrics for different training methods for YOLO.}
  \label{tab:detection_one}
  \vspace{-5mm}
\end{table}

\section{Results}
\label{sec:results}
\begin{figure*}
\vspace{2.5mm}
	\centering
	\begin{subfigure}[b]{0.24\linewidth}
		\includegraphics[width=\linewidth]{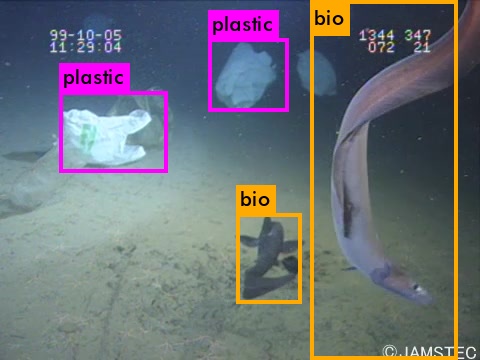}
		\caption{YOLOv2}
		\label{fig:yolo_several}	
	\end{subfigure}\hspace{1mm}
	\begin{subfigure}[b]{0.24\linewidth}
		\includegraphics[width=\linewidth]{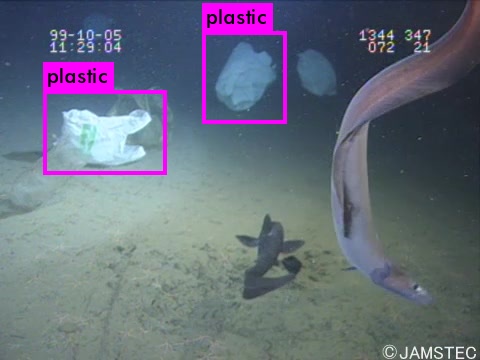}
		\caption{Tiny-YOLO}
		\label{fig:tiny_several}
	\end{subfigure}
    \begin{subfigure}[b]{0.24\linewidth}
		\includegraphics[width=\linewidth]{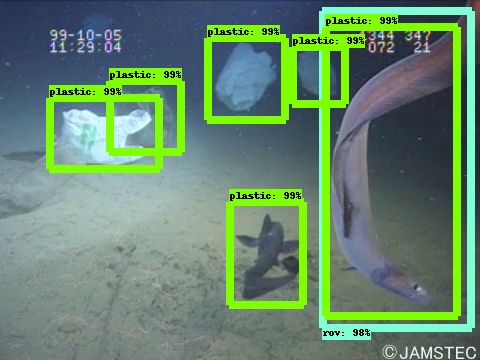}
		\caption{Faster R-CNN}
		\label{fig:rcnn_several}	
	\end{subfigure}\hspace{1mm}
	\begin{subfigure}[b]{0.24\linewidth}
		\includegraphics[width=\linewidth]{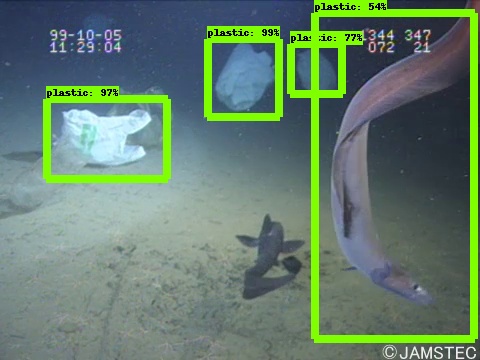}
		\caption{SSD}
		\label{fig:ssd_several}
	\end{subfigure}
    
    \begin{subfigure}[b]{0.24\linewidth}
		\includegraphics[width=\linewidth]{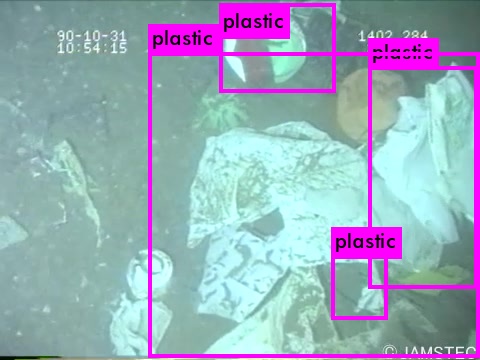}
		\caption{YOLOv2}
		\label{fig:yolo_many}	
	\end{subfigure}\hspace{1mm}
	\begin{subfigure}[b]{0.24\linewidth}
		\includegraphics[width=\linewidth]{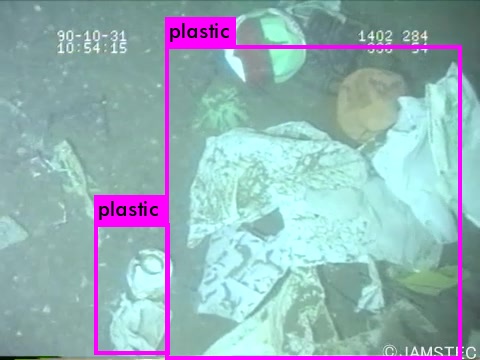}
		\caption{Tiny-YOLO}
		\label{fig:tiny_many}
	\end{subfigure}
    \begin{subfigure}[b]{0.24\linewidth}
		\includegraphics[width=\linewidth]{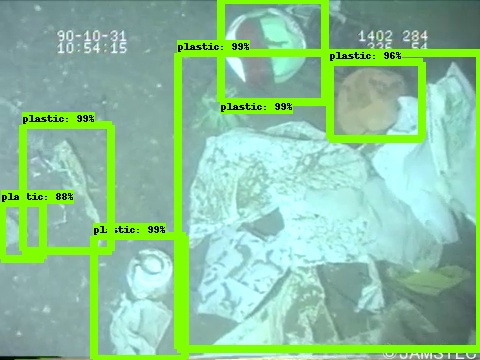}
		\caption{Faster R-CNN}
		\label{fig:rcnn_many}	
	\end{subfigure}\hspace{1mm}
	\begin{subfigure}[b]{0.24\linewidth}
		\includegraphics[width=\linewidth]{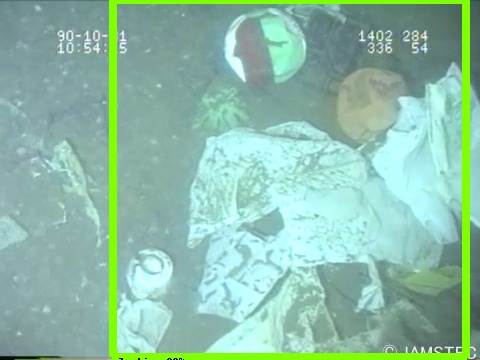}
		\caption{SSD}
		\label{fig:ssd_many}
	\end{subfigure}
    
    
    \begin{subfigure}[b]{0.24\linewidth}
		\includegraphics[width=\linewidth]{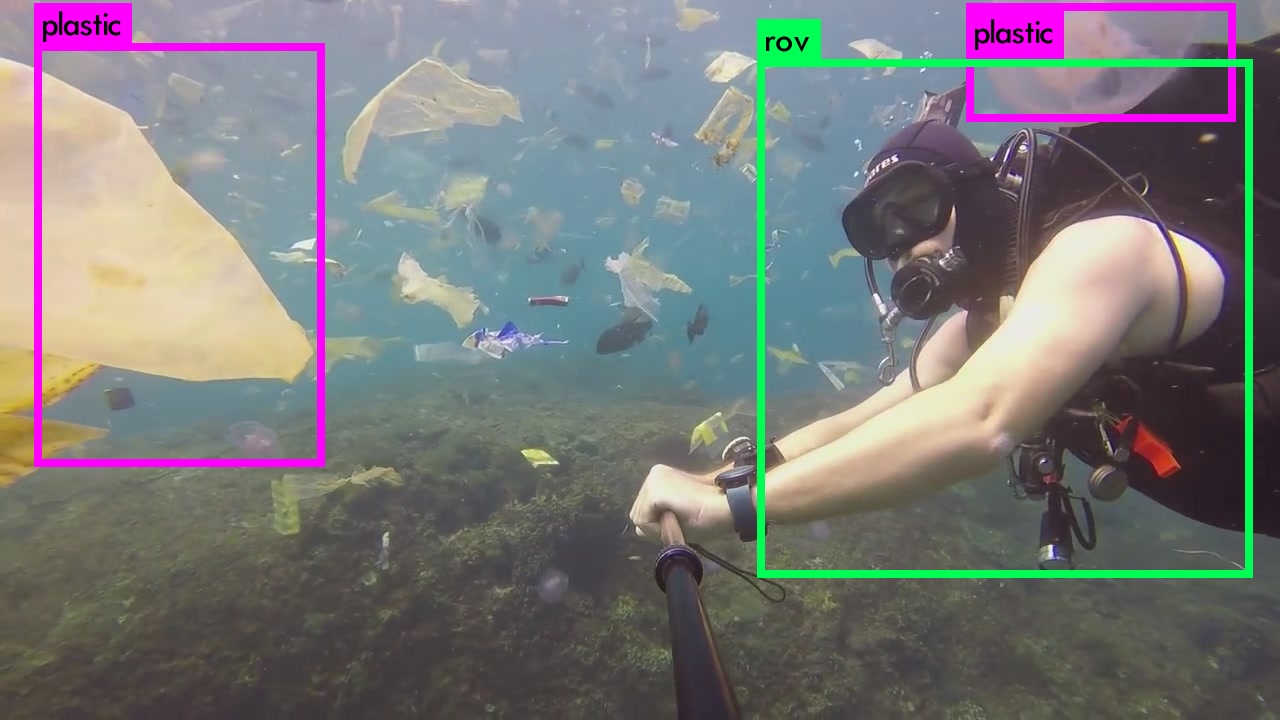}
		\caption{YOLOv2}
		\label{fig:yolo_bali}	
	\end{subfigure}\hspace{1mm}
	\begin{subfigure}[b]{0.24\linewidth}
		\includegraphics[width=\linewidth]{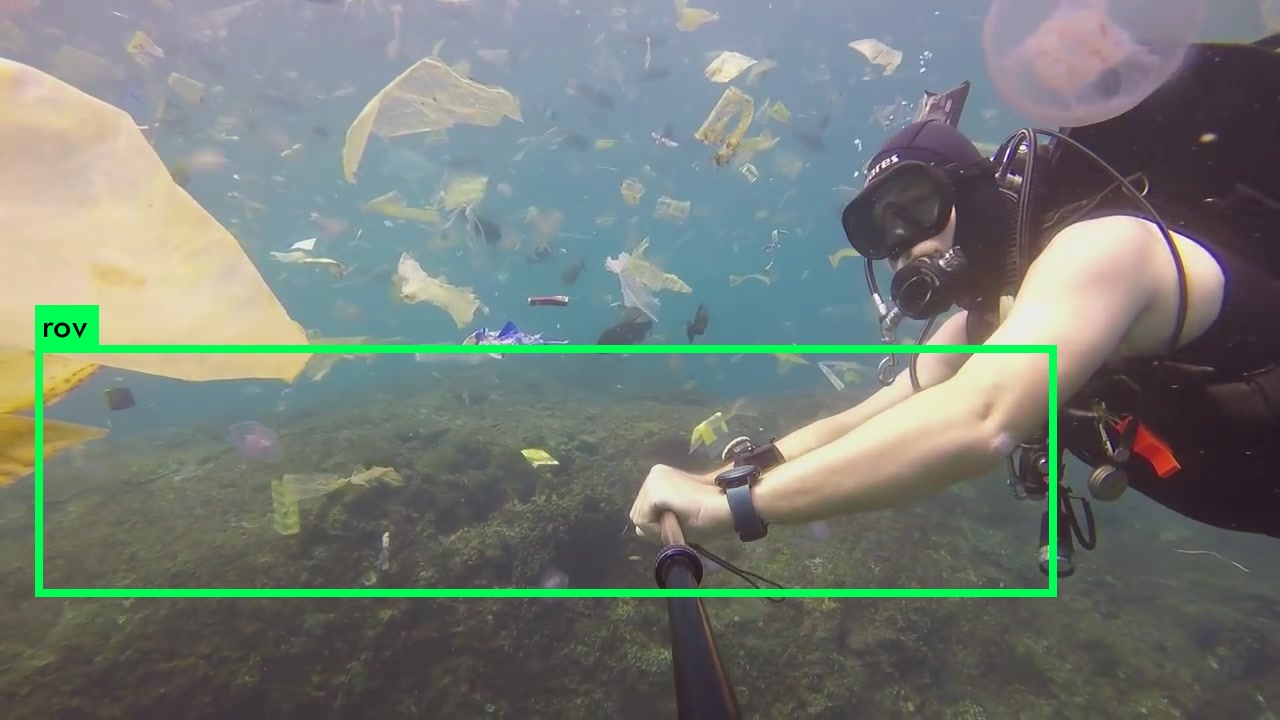}
		\caption{Tiny-YOLO}
		\label{fig:tiny_bali}
	\end{subfigure}
    \begin{subfigure}[b]{0.24\linewidth}
		\includegraphics[width=\linewidth]{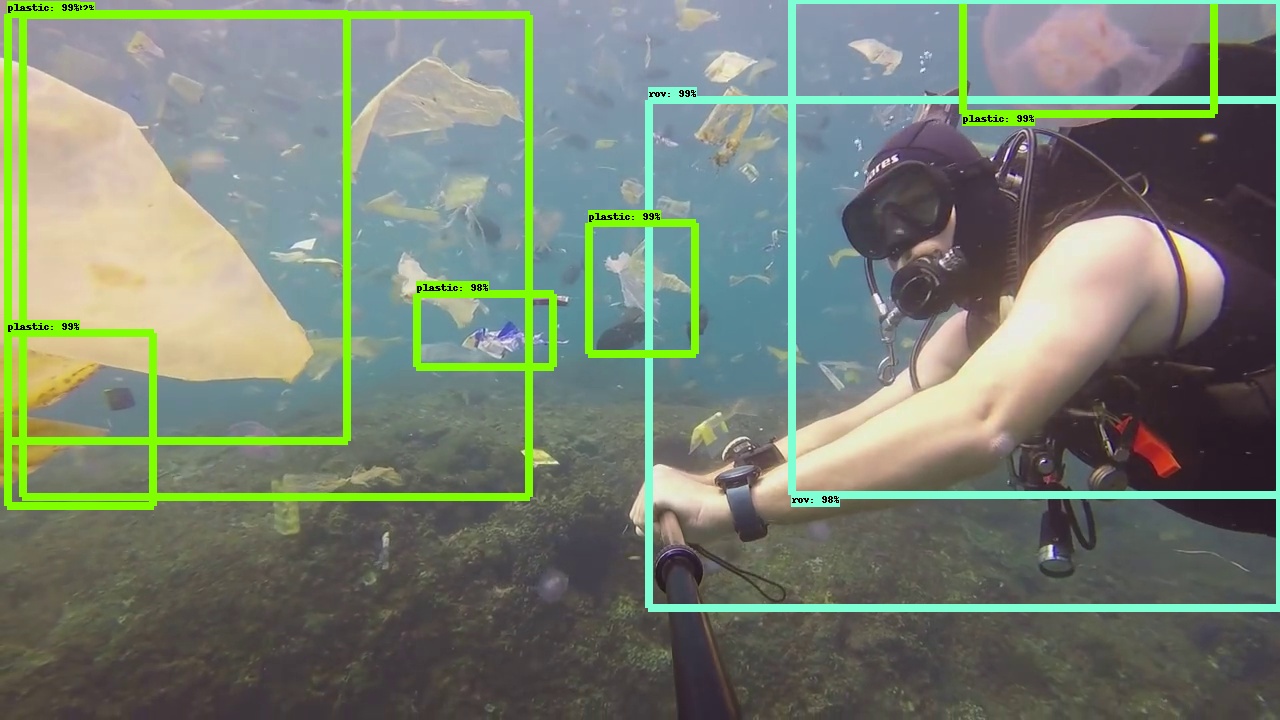}
		\caption{Faster R-CNN}
		\label{fig:rcnn_bali}	
	\end{subfigure}\hspace{1mm}
	\begin{subfigure}[b]{0.24\linewidth}
		\includegraphics[width=\linewidth]{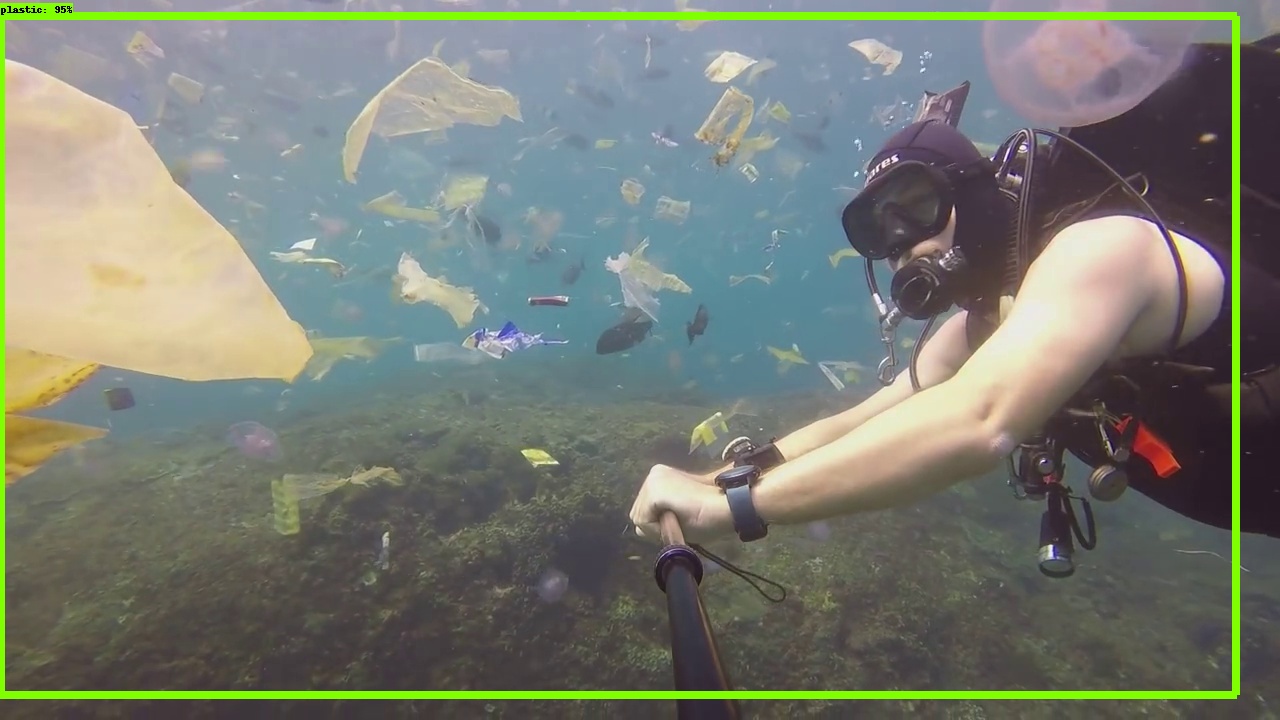}
		\caption{SSD}
		\label{fig:ssd_bali}
	\end{subfigure}
    \caption{Detection results on two images from test data for all networks. Images \ref{fig:yolo_bali}-\ref{fig:ssd_bali} are from \cite{noauthor_ocean_nodate}}
    \label{fig:bigolefigure}
\end{figure*}

The results below were obtained on the test dataset described in subsection \ref{subsec:evalData}, which was designed to be challenging to the models and to provide a true to life representation of how a marine debris detector would have to operate. The videos from which the test data is sourced are mostly from 10 years or more before the time of videos in the training set; they differ from each other and the training set in location, depth, and ROV. Because of these factors, the test set provides a good indicator of how these detectors would perform across different platforms over the course of many years and through many environmental changes. This does degrade the performance of the detector from what could have been achieved if more similar data was chosen, but it is a better evaluation of what the detector's performance would be in the field. Along with evaluating the data on our test set, we also processed three additional videos, with each network, which can be seen in Figure \ref{fig:bigolefigure} as well as in the video submitted with this paper. The third video, which is shown in Figures \ref{fig:yolo_bali} - \ref{fig:ssd_bali}, is a viral video of ocean trash recorded in Bali ~\cite{noauthor_ocean_nodate}. Being from an entirely different visual environment than our training and test data, it challenges all of our detectors greatly, but some promising detections can be seen over the course of the video.

\subsection{Quantitative Results}
The results shown in Table \ref{tab:detection_one} display some known traits of the network architectures used. Overall, YOLOv2 and Tiny-YOLO have lower mAP when compared to Faster R-CNN and SSD. Conversely, Faster R-CNN and SSD have higher processing times, as seen in Table III. These traits are well known, but it is important to note that they remain consistent with other work in this application. This trade-off between mAP and FPS does not affect IoU, however. All four network architectures have similar IoU values, meaning that none are the clear victor in terms of how accurate their bounding boxes are. 

In terms of which method would be the ideal for underwater trash detection, Faster R-CNN is the obvious choice purely from a standpoint of accuracy, but falters in terms of inference time.  YOLOV2 strikes a good balance of accuracy and speed, while SSD provides the best inference times on CPU. If performance is the primary consideration, however, Tiny-YOLO outpaces all other algorithms significantly on the TX2, the most realistic hardware for a modern AUV.  

We can also see the results of including classes with too few examples, particularly in the case of the bio class, which has many fewer examples in the training set, despite being at least as varied as the plastic trash in the underwater environment. The fact that classes with few training images relative to their complexity will perform well is not surprising in any way, but highlights some of the difficulties inherent in creating a dataset for trash detection: some objects may only be encountered a handful of times but should still count as trash, or be marked as biological and therefore not be removed. 

It is worth noting that in the case of the bio class, the versions of YOLO trained using transfer learning greatly increase accuracy. We believe this to be related to the types of objects that the pre-trained weights were trained to detect and the preponderance of plastic objects in the dataset. Simply put, by not updating those earlier layers, we avoid skewing the basic image features towards plastic objects, reducing plastic detection accuracy slightly which increasing that of the bio class.

\subsection{Qualitative Results}
The most substantial result from this evaluation is the answer to the question we set out to investigate: can we use deep-learning based visual object detection methods to detect marine debris in real-time? The answer, borne out by our detection accuracy and performance results is, yes, plausibly. There are some difficulties in constructing a representative dataset, but overall our detection results lead us to believe that detectors similar to these could be used to find trash underwater with high enough accuracy to be useful. Moreover, the observed performance, especially on the NVIDIA\texttrademark{} Jetson TX2, is very encouraging for the prospect of a real-time application. The Jetson is small enough to conceivably be used in an AUV, and although it increases the electrical and thermal load of the system, the authors believe the increase to be manageable, with the Jeston only requiring $15$ watts \cite{jetsontx2} and producing minimal heat. This, combined with its performance in tasks such as these, makes the Jetson ideal for mobile robotics, even for our specific use case. With a Jetson installed, any AUV should be able to achieve real-time trash detections with reasonable accuracy. The authors' Aqua AUV \cite{Sattar08IROS} will be fitted with a TX2 in the near future for this very purpose.

\section{FUTURE WORK}
In the future, we hope to expand upon this work by testing similar algorithms on a dataset collected from our own observations of marine debris in real-world environments.Additionally, we plan to explore other approaches to this task. The problem of detecting trash in underwater environments causes some inherent difficulties in deep-learning-based solutions, as it is difficult, verging on impossible, to collect a dataset with sufficient coverage. The use of transfer learning, progressive neural networks, or other methods which help to overcome data limitations are an area we intend to explore. In a similar vein, we hope to explore generative methods for data synthesis to help increase the amount of available trash data for training.

Of course, our primary future direction is the general goal of this work: to move towards an AUV capable of clearing marine environments of trash, with little to no human help. This will involve the development of navigation, exploration, and manipulation strategies to better find all of the trash in the water and remove it for good. We are very interested in the possibility of multi-robot collaboration between multi-modal robots (for instance aerial and aquatic) in solving this problem.

\section{Conclusion}
\label{sec:conclusion}
In this paper, we applied and evaluated four deep-learning-based object detectors to the problem of finding marine debris, particularly plastic, in an attempt to answer the question "Can visual learning object detectors be used to detect trash underwater in real-time?" We built a dataset from publicly available data and devised a data model with single class for all trash detections. Following this, we trained YOLOv2, Tiny-YOLO, Faster R-CNN, and SSD using standard fine-turning procedures, then evaluated them on a test dataset composed of challenging objects that had not been seen during training. We also trained YOLOv2 using transfer learning techniques to train only the last four and the last three layers. We performed evaluations on three different devices, acquiring information about the mAP and IoU for each model and network, as well as runtime speeds for all configurations. Based on these results, we feel that trash detection using visual deep learning models is plausible in real-time. In the future, these techniques should be used as the first part of an AUV designed to remove trash from aquatic environments which will be a critical tool in our fight against the pollution of our oceans. 

\section*{Acknowledgments}
We gratefully acknowledge the support of the MnDRIVE initiative and the support of NVIDIA Corporation with the donation of the Titan Xp GPU used for this research.
The dataset used in this work was created based off the publicly available J-EDI dataset, provided by JAMSTEC, the Japanese Agency for Marine-Earth Science and Technology. The authors wish to thank them for providing this data for scientific endeavors. 
The authors also wish to thank Sophie Fulton, Marc Ho, Elliott Imhoff, Yuanzhe Liu, Julian Lagman, and Youya Xia for their tireless efforts in annotating images for training purposes.
\bibliographystyle{abbrv}
\bibliography{allbibs}

\begin{thebibliography}{10}

\bibitem{alexeyab}
AlexeyAB.
\newblock Yolo: How to train (to detect your custom objects).
\newblock
  \url{https://github.com/AlexeyAB/darknet#how-to-train-to-detect-your-custom-objects)},
  2018.
\newblock Accessed: 9-14-2018.

\bibitem{Bernstein2013IROS}
M.~Bernstein, R.~Graham, D.~Cline, J.~M. Dolan, and K.~Rajan.
\newblock Learning-based event response for marine robotics.
\newblock In {\em 2013 IEEE/RSJ International Conference on Intelligent Robots
  and Systems}, pages 3362--3367, Nov 2013.

\bibitem{derraik2002pollution}
J.~G. Derraik.
\newblock The pollution of the marine environment by plastic debris: a review.
\newblock {\em Marine pollution bulletin}, 44(9):842--852, 2002.

\bibitem{fabbri_ehancing_2018}
C.~Fabbri, M.~J. Islam, and J.~Sattar.
\newblock Enhancing underwater imagery using generative adversarial networks.
\newblock In {\em 2018 IEEE International Conference on Robotics and Automation
  (ICRA)}, pages 7159--7165, May 2018.

\bibitem{jetsontx2}
D.~Franklin.
\newblock {\em NVIDIA Jetson TX2 Delivers Twice the Intelligence to the Edge}.
\newblock {\small
  \url{https://devblogs.nvidia.com/jetson-tx2-delivers-twice-intelligence-edge/}}.
  Accessed 03-01-2018.

\bibitem{ge_semi-automatic_2016}
Z.~Ge, H.~Shi, X.~Mei, Z.~Dai, and D.~Li.
\newblock Semi-automatic recognition of marine debris on beaches.
\newblock {\em Scientific Reports}, 6:25759, May 2016.

\bibitem{Girdhar2011MARE}
Y.~Girdhar, A.~Xu, B.~B. Dey, M.~Meghjani, F.~Shkurti, I.~Rekleitis, and
  G.~Dudek.
\newblock {MARE: Marine Autonomous Robotic Explorer}.
\newblock In {\em Proceedings of the 2011 IEEE/RSJ International Conference on
  Intelligent Robots and Systems (IROS '11)}, pages 5048--5053, San Francisco,
  USA, September 2011.

\bibitem{Girshick2014RCNN_CVPR}
R.~Girshick, J.~Donahue, T.~Darrell, and J.~Malik.
\newblock {Rich Feature Hierarchies for Accurate Object Detection and Semantic
  Segmentation}.
\newblock In {\em Proceedings of the 2014 IEEE Conference on Computer Vision
  and Pattern Recognition}, CVPR '14, pages 580--587, Washington, DC, USA,
  2014. IEEE Computer Society.

\bibitem{hidalgo2015review}
F.~Hidalgo and T.~Br{\"a}unl.
\newblock {Review of underwater SLAM techniques}.
\newblock In {\em Automation, Robotics and Applications (ICARA), 2015 6th
  International Conference on}, pages 306--311. IEEE, 2015.

\bibitem{noauthor_ocean_nodate}
R.~Horner.
\newblock The ocean currents brought us in a lovely gift today... - {YouTube}.

\bibitem{howell_north_2012}
E.~A. Howell, S.~J. Bograd, C.~Morishige, M.~P. Seki, and J.~J. Polovina.
\newblock On {North} {Pacific} circulation and associated marine debris
  concentration.
\newblock {\em Marine Pollution Bulletin}, 65(1):16--22, Jan. 2012.

\bibitem{JAMSTECDebri}
{\relax Japan Agency for Marine Earth Science and Technology}.
\newblock {\em Deep-sea Debris Database}.
\newblock {\small
  \url{http://www.godac.jamstec.go.jp/catalog/dsdebris/e/index.html}}. Accessed
  02-10-2018.

\bibitem{kim2015active}
A.~Kim and R.~M. Eustice.
\newblock {Active Visual SLAM for Robotic Area Coverage: Theory and
  Experiment}.
\newblock {\em The International Journal of Robotics Research},
  34(4-5):457--475, 2015.

\bibitem{Kulkarni2013CARE}
S.~Kulkarni and S.~Junghare.
\newblock Robot based indoor autonomous trash detection algorithm using
  ultrasonic sensors.
\newblock In {\em 2013 International Conference on Control, Automation,
  Robotics and Embedded Systems (CARE)}, pages 1--5, Dec 2013.

\bibitem{leonard2016autonomous}
J.~J. Leonard and A.~Bahr.
\newblock {Autonomous underwater vehicle navigation}.
\newblock In {\em Springer Handbook of Ocean Engineering}, pages 341--358.
  Springer, 2016.

\bibitem{liu2016ssd}
W.~Liu, D.~Anguelov, D.~Erhan, C.~Szegedy, S.~Reed, C.-Y. Fu, and A.~C. Berg.
\newblock {SSD: Single shot multibox detector}.
\newblock In {\em European conference on computer vision}, pages 21--37.
  Springer, 2016.

\bibitem{lu_underwaterimage_2016}
H.~Lu, Y.~Li, X.~Xu, L.~He, Y.~Li, D.~Dansereau, and S.~Serikawa.
\newblock Underwater image descattering and quality assessment.
\newblock In {\em 2016 IEEE International Conference on Image Processing
  (ICIP)}, pages 1998--2002, Sept 2016.

\bibitem{mace_at-sea_2012}
T.~H. Mace.
\newblock At-sea detection of marine debris: {Overview} of technologies,
  processes, issues, and options.
\newblock {\em Marine Pollution Bulletin}, 65(1):23--27, Jan 2012.

\bibitem{rozaliaROV}
R.~Z. Miller.
\newblock {\em {Underwater Robots Clean Up Marine Debris on the Seafloor}}.
\newblock The Rozalia Project.
\newblock {\small
  \url{https://www.sea-technology.com/features/2012/1212/underwater_robots.php}}.
  Accessed 02-10-2018.

\bibitem{marinedebris}
{\relax National Oceanic and Atmospheric Administration}.
\newblock {\em What is marine debris?}
\newblock {\small \url{https://oceanservice.noaa.gov/facts/marinedebris.html}}.
  Accessed 02-10-2018.

\bibitem{dianna.parker_detecting_2015}
{\relax National Oceanic and Atmospheric Administration}.
\newblock Detecting {Japan} {Tsunami} {Marine} {Debris} at {Sea}: {A}
  {Synthesis} of {Efforts} and {Lessons}-{Learned} {\textbar} {OR}\&{R}'s
  {Marine} {Debris} {Program}, Jan 2015.

\bibitem{clearBlueSea}
{\relax Non-Profit Organization}.
\newblock {\em {Clear Blue Sea}}.
\newblock {\small \url{http://clearbluesea.org}}. Accessed 02-10-2018.

\bibitem{paull2014auv}
L.~Paull, S.~Saeedi, M.~Seto, and H.~Li.
\newblock {AUV navigation and localization: A review}.
\newblock {\em IEEE Journal of Oceanic Engineering}, 39(1):131--149, 2014.

\bibitem{redmon2016you}
J.~Redmon, S.~Divvala, R.~Girshick, and A.~Farhadi.
\newblock You only look once: Unified, real-time object detection.
\newblock In {\em Proceedings of the IEEE Conference on Computer Vision and
  Pattern Recognition}, pages 779--788, 2016.

\bibitem{redmon2016yolo9000}
J.~Redmon and A.~Farhadi.
\newblock {YOLO9000: Better, Faster, Stronger}.
\newblock {\em arXiv preprint arXiv:1612.08242}, 2016.

\bibitem{renNIPS15fasterrcnn}
S.~Ren, K.~He, R.~Girshick, and J.~Sun.
\newblock Faster {R-CNN}: Towards real-time object detection with region
  proposal networks.
\newblock In {\em Advances in Neural Information Processing Systems ({NIPS})},
  2015.

\bibitem{sandler2018inverted}
M.~Sandler, A.~Howard, M.~Zhu, A.~Zhmoginov, and L.-C. Chen.
\newblock Inverted residuals and linear bottlenecks: Mobile networks for
  classification, detection and segmentation.
\newblock {\em arXiv preprint arXiv:1801.04381}, 2018.

\bibitem{Sattar08IROS}
J.~Sattar, G.~Dudek, O.~Chiu, I.~Rekleitis, A.~Mills, P.~Gigu\`ere,
  N.~Plamondon, C.~Prahacs, Y.~Girdhar, M.~Nahon, and J.-P. Lobos.
\newblock Enabling autonomous capabilities in underwater robotics.
\newblock In {\em Proceedings of the {IEEE/RSJ} International Conference on
  Intelligent Robots and Systems ({IROS})}, pages 3628--3634, Nice, France,
  September 2008.

\bibitem{schlining_debris_2013}
K.~Schlining, S.~v. Thun, L.~Kuhnz, B.~Schlining, L.~Lundsten, N.~J. Stout,
  L.~Chaney, and J.~Connor.
\newblock Debris in the deep: {Using} a 22-year video annotation database to
  survey marine litter in {Monterey} {Canyon}, central {California}, {USA}.
\newblock {\em Deep Sea Research Part I: Oceanographic Research Papers}, 79:96
  -- 105, 2013.

\bibitem{simonyan2014very}
K.~Simonyan and A.~Zisserman.
\newblock Very deep convolutional networks for large-scale image recognition.
\newblock {\em arXiv preprint arXiv:1409.1556}, 2014.

\bibitem{stutters2008navigation}
L.~Stutters, H.~Liu, C.~Tiltman, and D.~J. Brown.
\newblock {Navigation technologies for autonomous underwater vehicles}.
\newblock {\em IEEE Transactions on Systems, Man, and Cybernetics, Part C
  (Applications and Reviews)}, 38(4):581--589, 2008.

\bibitem{szegedy2016rethinking}
C.~Szegedy, V.~Vanhoucke, S.~Ioffe, J.~Shlens, and Z.~Wojna.
\newblock Rethinking the inception architecture for computer vision.
\newblock In {\em Proceedings of the IEEE Conference on Computer Vision and
  Pattern Recognition}, pages 2818--2826, 2016.

\bibitem{tfzoo}
Tensorflow.
\newblock Tensorflow object detection zoo.
\newblock
  \url{https://github.com/tensorflow/models/blob/master/research/object_detection/g3doc/detection_model_zoo.md},
  2017.
\newblock Accessed: 2-20-2018.

\bibitem{epa1}
{\relax United States Environmental Protection Agency}.
\newblock {\em Sources of Aquatic Trash}.
\newblock {\small
  \url{https://www.epa.gov/trash-free-waters/sources-aquatic-trash}}. Accessed
  02-10-2018.

\bibitem{Toro2016Trash}
M.~Valdenegro-Toro.
\newblock Submerged marine debris detection with autonomous underwater
  vehicles.
\newblock In {\em 2016 International Conference on Robotics and Automation for
  Humanitarian Applications (RAHA)}, pages 1--7, Dec 2016.

\bibitem{s17051174}
X.~Yuan, J.-F. Martínez-Ortega, J.~A.~S. Fernández, and M.~Eckert.
\newblock {AEKF-SLAM: A New Algorithm for Robotic Underwater Navigation}.
\newblock {\em Sensors}, 17(5), 2017.

\end{thebibliography}
\end{document}